\documentclass[10pt]{llncs}
\usepackage{url}
\usepackage[utf8]{inputenc}
\usepackage{amsmath}
\usepackage{amssymb}
\usepackage{graphicx}
\usepackage{cite}
\usepackage{color, colortbl} 

\definecolor{Gray}{gray}{0.9}

\title{ Emotional Intensity analysis in Bipolar subjects}

\author{Carrillo, Facundo\inst{1} \and Mota, Natalia\inst{2} \and Copelli, Mauro\inst{3} \and Ribeiro, Sidarta\inst{2} \and Sigman, Mariano\inst{4} \and Cecchi, Guillermo\inst{5} \and Fernandez Slezak, Diego\inst{1}}

\institute{Laboratorio de Inteligencia Artificial Aplicada, Departamento de Computación\\
 Facultad de Ciencias Exactas y Naturales, Universidad de Buenos Aires, CABA, Argentina 
 \and
 Instituto do Cérebro, Universidade Federal do Rio Grande do Norte,  Natal, Brazil \\
 \and
 Universidade Federal de Pernambuco, Recife, Brazil \\
 \and
 Universidad Torcuato Di Tella, CABA, Argentina\\
 \and
 T.J. Watson Research Center, IBM, Yorktown Heights, NY, USA\\
 }

\begin{document}

\mainmatter              

\maketitle

\begin{abstract}
The massive availability of digital repositories of human thought opens radical novel way of studying the human mind.
Natural language processing tools and computational models have evolved such that many mental conditions are predicted by analysing speech.
Transcription of interviews and discourses are analyzed using syntactic, grammatical or sentiment analysis to infer the mental state.
Here we set to investigate if classification of Bipolar and control subjects is possible.
We develop the Emotion Intensity Index based on the Dictionary of Affect, and find that subjects categories are distinguishable.
Using classical classification techniques we get more than 75\% of labeling performance.
These results sumed to previous studies show that current automated speech analysis is capable of identifying altered mental states towards a quantitative psychiatry.

\end{abstract}

\section{Introduction}

Recent advances in technology allow scientists to examine the brain and the mind in different and radically novel ways. 
The virtually infinite repository of human computation in digital format allows the exploration of procedures in high dimensional spaces previously intractable with data generated in the laboratory. 

Psychology has historically used speech and conversation with patients as window to the mind. 
In the last years many computational techniques brought to the psychiatric-psychological practice many new analyical tools. 
Some mental conditions are prognosticated extremely well using only computational models based on speech. 
In the work \cite{mota2012speech}, the authors introduced a graphs model to characterize psychosis produced by schizophrenia and bipolar disorders. 
Authors showed how differences of topology of speech graphs may change depending on mental condition of the patients. 
This technique, based on the structure of graphs, captures properties related to thought organization and allows using the model as a predictor of mental disorders. 

Natural language processing of speech has gotten to predict psycosis onset better than the psychiatry experts \cite{bedi2015automated}. 
By analyzing coherence in speech, authors are able to predict the first psychotic episode in high-risk patients, identifying which patients are going to develop schizophrenia with a performance of 100\%. 
Moreover, computational models may detect mental alterations caused by drugs ingest \cite{bedi2014window}. 

In \cite{mota2014graph}, the authors ran an experiment where they ask the subjects (20 control, 20 bipolars -- 14 on mania and 6 on depression --, and 20 schizophrenic) to report a recent dream and report their waking activities immediately before that dream. 
Using these two different graphs they computed graph measures -- as features of each subject's speech --, and built a classification model to predict the group label of each subject: Control, Bipolar or schizophrenic. 
Binary classification (Control vs Bipolar) using waking reports is close to chance.
In particular, the worst classification was obtained for the maniac vs control comparison. 

Although the speech graphs are a great tool to characterize between schizophrenia and control or bipolar with dream and waking reports, they do not seem useful to classify between bipolar and control using waking reports, and the need to get a dream report from a psychotic patient can be really challenging. This may be because the lack of evidence of grammatical alterations in bipolar speech.

However, evidence shows that emotions change in subjects with mania disorder \cite{platman1969emotion,johnson2007emotion,gruber2008risk}. 
The result presented in \cite{strakowski2012functional} suggests, using fMRI, that abnormal modulation between ventral prefrontal and limbic regions, especially the amygdala, are likely contribute to poor emotional regulation and mood symptoms. 

We hipothesize that this poor emotional regulation and mood symptoms must be detected in speech, and thus the identification of patients should be possible analysing the emotional intensity in speech.
For this, we build a emotional intensity index to study the differences between control and maniac group using the same dataset of \cite{mota2014graph}.

\section{Methods}

\subsection{Subjects and interviews}
Forty subjects participated in the study, 20 of them were diagnosed as Bipolar and another 20 acted as control. The diagnosis was performed using the standard DSM IV ratings SCID\cite{first2012structured}. The  subjects were patients of the Hospital Onofre Lopes (UFRN) and Hospital Machado, Natal, Brazil. All subjects were interviewed with the following tasks: \textit{Please report a recent dream} and \emph{Please report your waking activities immediately before that dream}. Their discourse was recorded and a blind-conditioned experimenter transcribed the recordings.
All subjects signed an informed consent for this study, which was approved by the UFRN Research Ethics Committee (permit \#102/06-98244); 
For this study, we concatenated the speech for both questions as a unique text.
As texts were in Portuguese, text were translated into English using Google Translate.

\subsection{Emotional Intensity Analysis algorithm}

In this work, we presented a simple algorithm to measure the emotional intensity in language. For this we used a list of words with high emotional value. To define the emotional value we used the Dictionary of Affect in Language (English DAL) \cite{whissell1989dictionary}. English DAL is tool designed to measure the emotional meaning of words. In that work, the authors ran an experiment where ask people to rate some words in three categories: pleasantness, activation, and imagery. We used as high emotional words, those where \emph{pleasantness} where in the first 20\% and in the last 20\%.

From this, we defined the emotional intensity (EI) of a sentence as the rate of words in the sentences that are in high emotional list: 
\begin{displaymath}
EI(s) = \frac{\sum^N_{i=0} (dal_+(s_i)+dal_-(s_i))}{N} 
\end{displaymath}
where $N$ is the number of words in sentence $s$ and $dal_{\pm}(w)$ returns 1 if word $w$ belongs to positive or negative affective DAL word list.

For example, the sentence: 
\begin{center}
\emph{
 This is a \color{red} beautiful \color{black} day}
\end{center}
has 5 words but just one belongs to the high emotional list, so the emotional intensity  is $0.2$.

To evaluate a text, we split it into sentences and calculate the EI for every one. Then as a summarized measure we report the mean and the standar deviation of the serie.

We used scikit-learn \cite{pedregosa2011scikit}, a Python module for machine learning to perform all classifiers. The classifiers were initialized with the default parameters.

\section{Results}

We present a simple method to measure the emotional intensity in text. 
To test that emotional intensity index (EII) captures differences in emotional contents, we defined two corpus: 
\begin{itemize}
 \item Emotionally-neutral content: Wikipedia Articles. 
 \item Emotionally-intense content: selection of Poems.
\end{itemize}
For the poem corpus we used 75 randomly selected texts from \url{http://100.best-poems.net/}. 
For the Wikipedia Articles corpus, we got 100 random Wikipedia articles \footnote{https://en.wikipedia.org/wiki/Special:Random}. 

\begin{figure}[ht]
\centerline{\includegraphics[width=\linewidth]{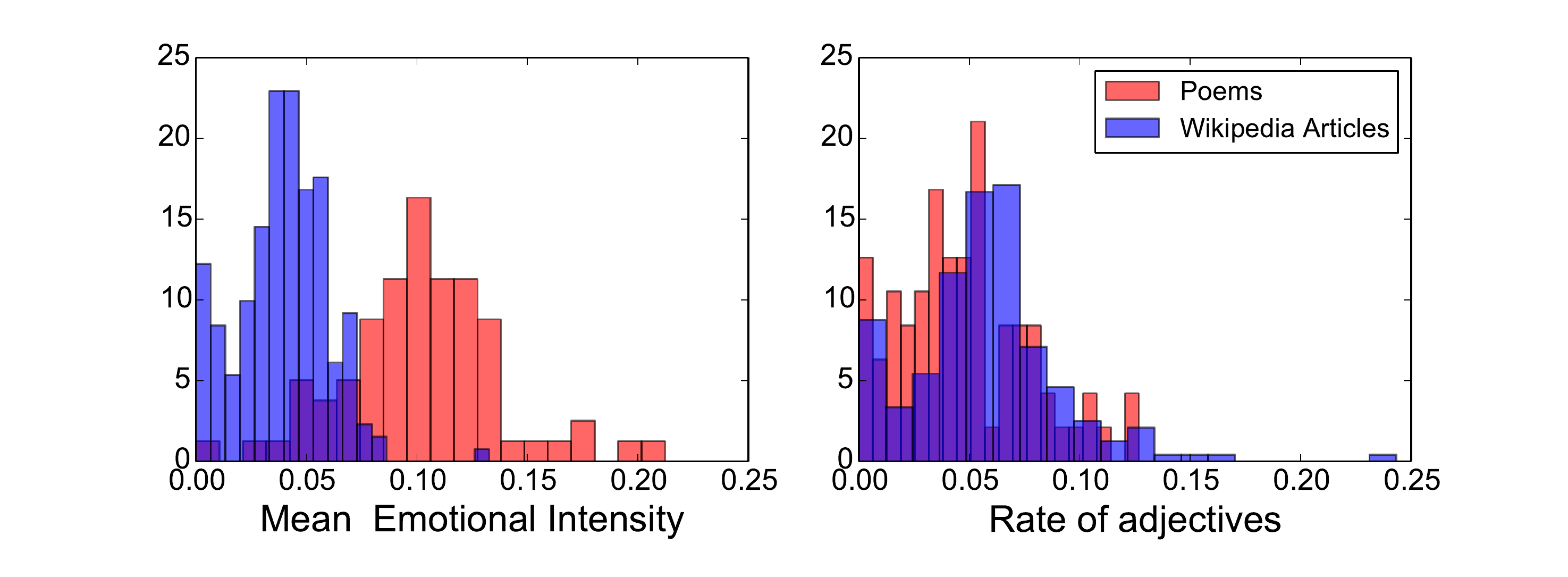}}
\caption{Emotional Intensity distribution and comparison to adjectives counting of intense emotional content (poems, red bars) and neutral content (wikipedia, blue bars).}
\label{fig:EM_validation}
\end{figure}

For each poem and each Wikipedia article, we estimated the EII (Figure \ref{fig:EM_validation}A). 
The Wikipedia Articles presented lower mean EII (0.0394 $\pm$ 0.0207) than documents in the poem corpus (0.1017 $\pm$ 0.0363).
Differences between both distributions were highly significant (ttest, $p \leq 10^{-46}$). 

To control we were not just counting the rate of adjective in the text we computed the part of speech (using NLTK \cite{loper2002nltk}) and compare the distribution of both corpus. Figure \ref{fig:EM_validation} B shows this comparison, here the distribution are not significantly distinct and presented a similar mean, for Wikipedia Articles corpus the adjective rate was $0.0574 \pm 0.0342$ and for the poems corpus the mean and standard deviation was $0.0473 \pm 0.0300$

The difference in the distribution using the emotional intensity and the not change in adjective rate shows that the emotional intensity algorithm are capturing the intensity of the speech and not only the use of adjectives that are mostly the vehicle of the modulation of the intensity in language.

In the introduction section we presented the hypothesis that we could observe change in the emotional intensity in maniac subjects. For this, we build the algorithm and showed before that it has an expected behavior in other validation corpus.

\begin{figure}[]
\centerline{\includegraphics[width=200px]{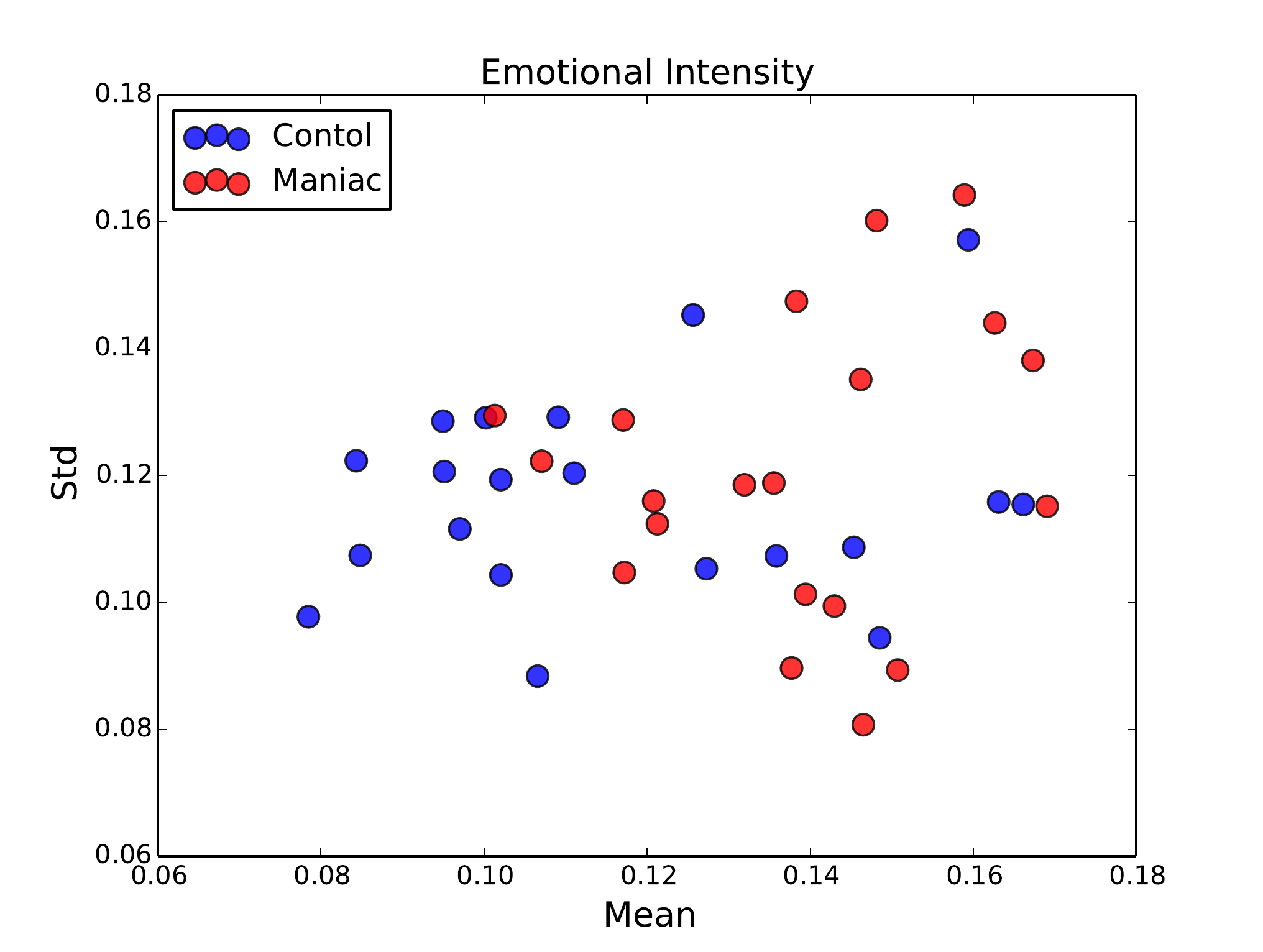}}
\caption{2D projection of subjects using mean and standard deviation as features.}
\label{fig:EI_maniac_vs_control}
\end{figure}

To address the main goal of this work, we measured the EI in 20 control subject and in 20 maniac subject. For each subject we measured the EI for every subject and then we reported the mean and the standard deviation. Figure \ref{fig:EI_maniac_vs_control} shows the organization in 2D of the subject if we used the mean and the standard deviation as features. The mean EI for the control subject are typically lower than the mean EI for  maniac subjects, it presented a statistically difference (pval=0.00793). The average and standard deviation for the mean EI in control group was 0.1168 $\pm$ 0.0277, and for maniac group was 0.1380 $\pm$ 0.0193.

Comparing the distrubtion of EI of control subjects and the previuos corpus we didn't find statistically difference between poems and control, they presented a very similar mean. However when we compared the maniac EI and poems we found that the first one not only had higher mean EI but they have presented statistically difference (pval =$4.211^{-05}$). This showed that maniac has even a higher EI than the poems dataset.

To analyze the emotional intensity as predictor model we ran a Logistic Regression using the mean EI as unique feature and a 10 fold cross-validation schema. With that configuration we obtained a performance of 75\% where chance was 50\%. 
We compared this classifier to other classical methods (see Table \ref{table:performances}), and Logistic Regression showed the best results.

 \begin{table}[]
\centering
\caption{Camparison of classification methods for labeling Control and Bipolar patients}
\label{table:performances}

\begin{tabular}{|c|c|c|c|c|c|c|}
\hline
\textbf{}                  & \multicolumn{2}{c|}{\textbf{Performance}} & \multicolumn{2}{c|}{\textbf{Roc Auc}} & \multicolumn{2}{c|}{\textbf{F1}} \\ \hline
\textbf{Classifier}        & \textbf{mean}        & \textbf{std}       & \textbf{mean}      & \textbf{std}     & \textbf{mean}   & \textbf{std}   \\ \hline
LogisticRegression         & 0.7500                & 0.2738           & 0.7000             & 0.4153         & 0.6166        & 0.4349       \\ \hline
LDA                        & 0.6750                & 0.2968           & 0.7000             & 0.4153         & 0.6066        & 0.4103       \\ \hline
SVC                        & 0.6500                & 0.3000           & 0.7000             & 0.4153         & 0.5766        & 0.4060       \\ \hline
GaussianNB                 & 0.6500                & 0.3000           & 0.6750             & 0.4038         & 0.5766        & 0.4060       \\ \hline
DecisionTreeClassifier     & 0.5500                & 0.2915           & 0.5250             & 0.2610         & 0.5533        & 0.3357       \\ \hline
GradientBoostingClassifier & 0.5750                & 0.2750           & 0.6375             & 0.2589         & 0.5266        & 0.3362       \\ \hline
BaggingClassifier          & 0.5750                & 0.1600           & 0.5875             & 0.2907         & 0.5066        & 0.2950       \\ \hline
KNeighborsClassifier       & 0.5500                & 0.1500           & 0.5500             & 0.1500         & 0.4300        & 0.2956       \\ \hline
RandomForestClassifier     & 0.5000                & 0.2500           & 0.6000             & 0.3570         & 0.4833        & 0.2833       \\ \hline
\end{tabular}
\end{table}
 
\section{Discussion}

Affective Computing has longly been trying to recognize and interpret human affect using computational techniques \cite{picard1995affective}.
These tecniques have evolved to Sentiment Analysis, where natural language processing tools -- combined with computational liguistics and classification methods -- recover subjective content from text\cite{pang2008opinion}.
Most applications are used to label, tag and analyze reviews and social media data.
These studies tipically use an emotion taxonomy, and try to quantify each emotion category, for example the six emotion proposed by Ekman \cite{ekman1993facial}.

We set to investigate if symptoms in Bipolar patients are identifiable in speech inferring the emotional intensity in their discourse.
We do not use a pre-defined emotion taxonomy and quantify each emotional axis. 
Instead, we propose a simpler approach where we just want to identify intense emotional content (in any category of emotion) and explore distributional differences in discourse from Control and Bipolar subjects.
For this purpose, we rely on DAL\cite{whissell1989dictionary}.

Our results showed that differences in emotional intensity may be detected using this simple method for different corpus of data.
However, this simple method is capturing more than just counting words, as counting adjectives did not separate between groups.

The application of this index was able to classify between control and bipolar subjects with a fair performance.
These results complement to previous studies where natural language processing tools are used to classify between schizophrenia and control patient\cite{mota2014graph,bedi2015automated}, towards quantitative psychiatry.

\section*{Acknowledgments}

This research was supported by University of Buenos Aires, CONICET (Argentina) and ANPCyT (Argentina). MS is sponsored by James McDonnell Foundation 21st Century Science Initiative in Understanding Human Cognition. DFS is sponsored by Microsoft Faculty Fellowship. This research is partly funded by the STIC-AmSud program partners MINCyT (Argentina), Inria (France), CAPES (Brasil) and ANII (Uruguay), through the COMPLEX project.

\end{document}